\documentclass[letterpaper, 10 pt, conference]{ieeeconf}  

\IEEEoverridecommandlockouts                              

\usepackage{latexsym}
\usepackage{amsmath}
\usepackage{amssymb}
\usepackage{epsfig}
\usepackage{bbm}
\usepackage{caption}
\usepackage{graphicx,float}
\usepackage{cuted}

\usepackage{cite}

\usepackage[ruled]{algorithm2e}

\usepackage{verbatim}
\usepackage{hyperref}
\hypersetup{%
  breaklinks,
  bookmarksnumbered=true,
  bookmarksopen=true,
  colorlinks=true,
  linkcolor=black,
  urlcolor=black,
  citecolor=black
}
\usepackage{xcolor}
\usepackage{subfiles}

\begin{document}
\bstctlcite{IEEEexample:BSTcontrol}
%

\title{TartanDrive 2.0: More Modalities and Better Infrastructure to Further Self-Supervised Learning Research in Off-Road Driving Tasks}








\author{Matthew Sivaprakasam$^{1}$, Parv Maheshwari$^{2}$, Mateo Guaman Castro$^{1}$, Samuel Triest$^{1}$, Micah Nye$^{3}$,\\ Steve Willits$^{1}$, Andrew Saba$^{1}$, Wenshan Wang$^{1}$, and Sebastian Scherer$^{1}$ 
\thanks{* This work was supported by ARL awards \#W911NF1820218 and \#W911NF20S0005.}%
\thanks{$^{1}$ Robotics Institute, Carnegie Mellon University, msivapra,mguamanc,striest,wenshanw,basti@andrew.cmu.edu}%
\thanks{$^{2}$ Department of Mathematics, Indian Institute of Technology Kharagpur. parvmaheshwari2002@iitkgp.ac.in}%
\thanks{$^{3}$ Department of Mechanical Engineering, University of Pittsburgh. man172@pitt.edu}%
}


\maketitle
\begin{strip}
  \centering
  \vspace*{-27mm}
  \includegraphics[width=\linewidth]{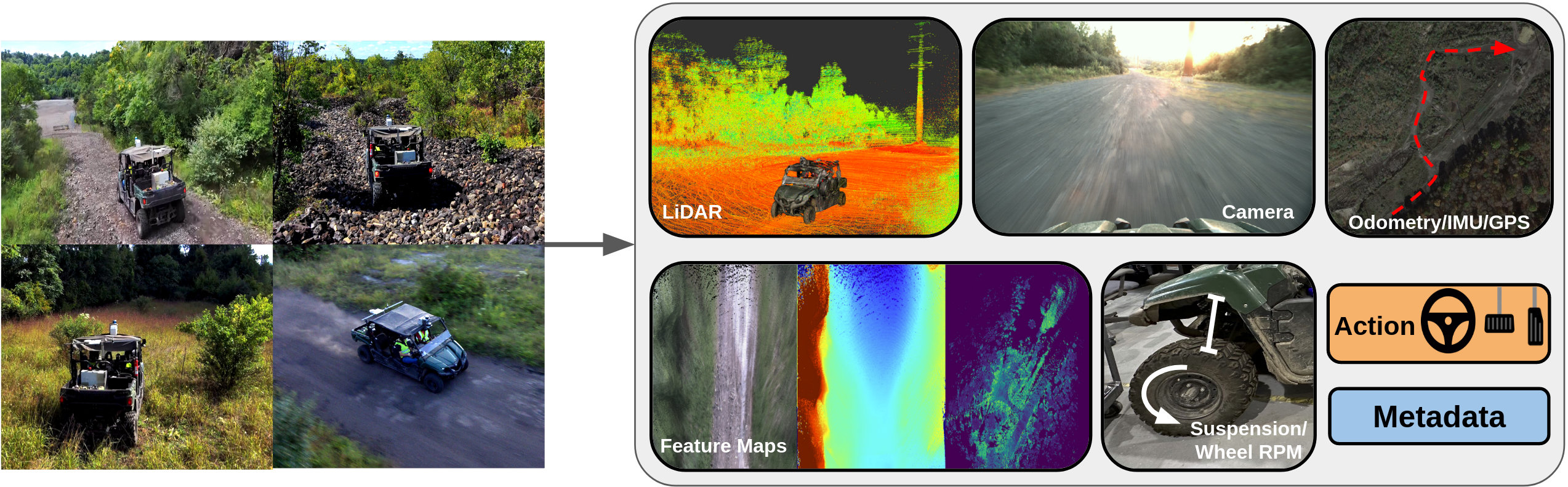}
  \captionof{figure}{We provide a new off-road driving dataset for self-supervised learning tasks. Multi-modal data is collected as the robot is driven through several types of terrain that present challenging scenarios to perception, planning, and control algorithms.}
  \label{fig:title}
  \vspace*{-1mm}
\end{strip}

\IEEEpeerreviewmaketitle

\begin{abstract}
We present TartanDrive 2.0, a large-scale off-road driving dataset for self-supervised learning tasks. In 2021 we released TartanDrive 1.0, which is one of the largest datasets for off-road terrain. As a follow-up to our original dataset, we collected seven hours of data at speeds of up to 15m/s with the addition of three new LiDAR sensors alongside the original camera, inertial, GPS, and proprioceptive sensors. We also release the tools we use for collecting, processing, and querying the data, including our metadata system designed to further the utility of our data. Custom infrastructure allows end users to reconfigure the data to cater to their own platforms. These tools and infrastructure alongside the dataset are useful for a variety of tasks in the field of off-road autonomy and, by releasing them, we encourage collaborative data aggregation. These resources lower the barrier to entry to utilizing large-scale datasets, thereby helping facilitate the advancement of robotics in areas such as self-supervised learning, multi-modal perception, inverse reinforcement learning, and representation learning. The dataset is available at \href{https://theairlab.org/TartanDrive2}{https://theairlab.org/TartanDrive2}.

\end{abstract}


\section{Introduction}

As the state of the art in autonomous driving improves, robots are expected to handle increasingly complex tasks. In off-road driving, this translates to situations such as a robot knowing whether the dark brown path in front of it is dry dirt or thick mud, whether it is worth it to take a shortcut through tall grass, and when to preemptively increase velocity in order to make it up a hill. Navigating these decisions requires behaviors that are extremely difficult to robustly design and tune by hand. Many of the currently-researched solutions to this challenge involve the adoption of large neural networks and data-driven models. For the task of on-road driving in urban scenarios, there are several large datasets available \cite{waymo_perception, waymo_motion, nuscenes, nuscenes_panoptic}. These datasets were feasible not only because of the resources present at the institutions that collected them, but also because of the inherent everyday nature of on-road driving. There exist fewer datasets for off-road driving, primarily because the nature of off-road terrain makes gathering enough data uniquely difficult. Collecting samples in simulation is often insufficient due to the complexity of calculating the dynamics of terrain in complex environments. In fact, most real-world scenarios that are hard to accurately simulate are the same ones which produce complex situations for autonomous off-road agents, such as dense foliage and slippery surfaces. However, collecting data in real-life presents logistics challenges such as preserving driver safety and dealing with frequent vehicle damage and wear. These difficulties have resulted in a scarcity of available data for off-road driving. 

While the number of off-road datasets has been growing \cite{9811648, jiang2020rellis3d, RUGD2019IROS, cavs, maturana, verti-wheelers, knights2023wildplaces, reinke2022iros}, many of them are still limited in sample size, modality, or difficulty \cite{bevnet}. This is often due to their focus on specific tasks such as semantic segmentation, which requires manual labeling effort. Scaling up these datasets would come either at the cost of time or label quality. It is possible to use multiple datasets together to train a model, but the variability of off-road terrain often causes inconsistency in labels across datasets \cite{DBLP}. The difficulty of generating large amounts of manually labeled data in off-road terrain suggests that self-supervised learning is essential to outperforming prior methods. Many works have shown that self-supervised methods can be used to learn strong representations \cite{NEURIPS2020_f3ada80d,oquab2023dinov2,pmlr-v119-chen20j}, and some works have already shown how it can be used for various tasks in off-road driving environments \cite{meng2023terrainnet,chen2023learningonthedrive,castro2023does,triest2023learning}. We argue that in order to scale up the amount of data for off-road tasks, datasets should be designed to be task-agnostic and with self-supervised learning in mind.

In 2021, we released TartanDrive, a multi-modal off-road driving dataset designed for dynamics modeling \cite{9811648}. In this work we present TartanDrive 2.0, a larger dataset geared towards self-supervised learning methods. The contributions of this dataset over the previous generation are as follows:
\begin{enumerate}
\item \textbf{More sensors and data:} In our original work, we argue that multiple modalities are essential for learning models for off-road terrain. To that end, we add LiDAR as an additional modality. We also collect seven hours over the five hours of data present in the original dataset, at higher average speeds and including some new areas, as well as areas present in the previous dataset that have dramatically changed over the past two years.
\item \textbf{Better infrastructure for end-users:} We have improved the infrastructure in order to improve ease-of-use and utility of the dataset, such as the ability to re-process the data based on user-specified configurations. We also implement a metadata system that filters and groups data by a variety of properties.
\item \textbf{Open-source data collection tooling:} Our data collection process has been constructed in a way that allows us to continually release more data over time as the seasons change and as we add more modalities. By releasing our tools and framework, others can collect their data and easily merge it with our own to create larger datasets.
\end{enumerate}

\section{Related Work}

Collecting general driving datasets is a common pursuit, the largest ones coming from companies gathering data in urban scenarios. For example, nuScenes by Motional is a dataset containing 1.4 million images and 390,000 LiDAR sweeps, with corresponding 3D object annotations \cite{nuscenes, nuscenes_panoptic}. The Waymo Open Dataset is another example with over a million images and LiDAR pointclouds \cite{waymo_perception, waymo_motion}. Both datasets include infrastructure and tools to query and process these massive amounts of data in ways that support various downstream learning tasks and benchmarks that are not often seen in smaller datasets. There are also other efforts in collecting data for on-road environments, such as the RACECAR dataset that includes raw and processed sensor data from autonomous racecars driving at high speeds \cite{kulkarni2023racecar}.

A number of earlier works have tried to circumvent the issues due to the lack of real-world off-road driving data by utilizing simulation environments. Tremblay et al. show how training a neural network on multiple modalities in simulation can allow a robot to predict its dynamics in unseen real-world data \cite{tremblay}. Sivaprakasam et al. collect data in simulation of a robot driving over obstacles in order to train a model that predicts the difference between a desired path and its resulting path \cite{matthew}.

Now, more off-road datasets have been collected. RUGD contains over 7,000 images in a variety of off-road terrain with manually-labeled semantic segmentation masks \cite{RUGD2019IROS}. Rellis-3D includes annotations for 6,235 images and 13,556 scans from two different LiDARs, as well as the bags that they originally recorded which also contain IMU, GPS, and stereo image data. Sharma et al. created an off-road image dataset where, rather than creating semantic masks, they label regions in an image based on their traversability by different types of vehicles \cite{cavs}. The first version of TartanDrive has also been available for two years and has been used for a number of tasks even outside the domain of off-road driving. Shah et al. have used this data as part of a bigger dataset that was used to train a large foundation model designed with vision-based robotic navigation in mind \cite{shah2023vint}.


\begin{figure}
    \centering
    \includegraphics[width=1.00\linewidth]{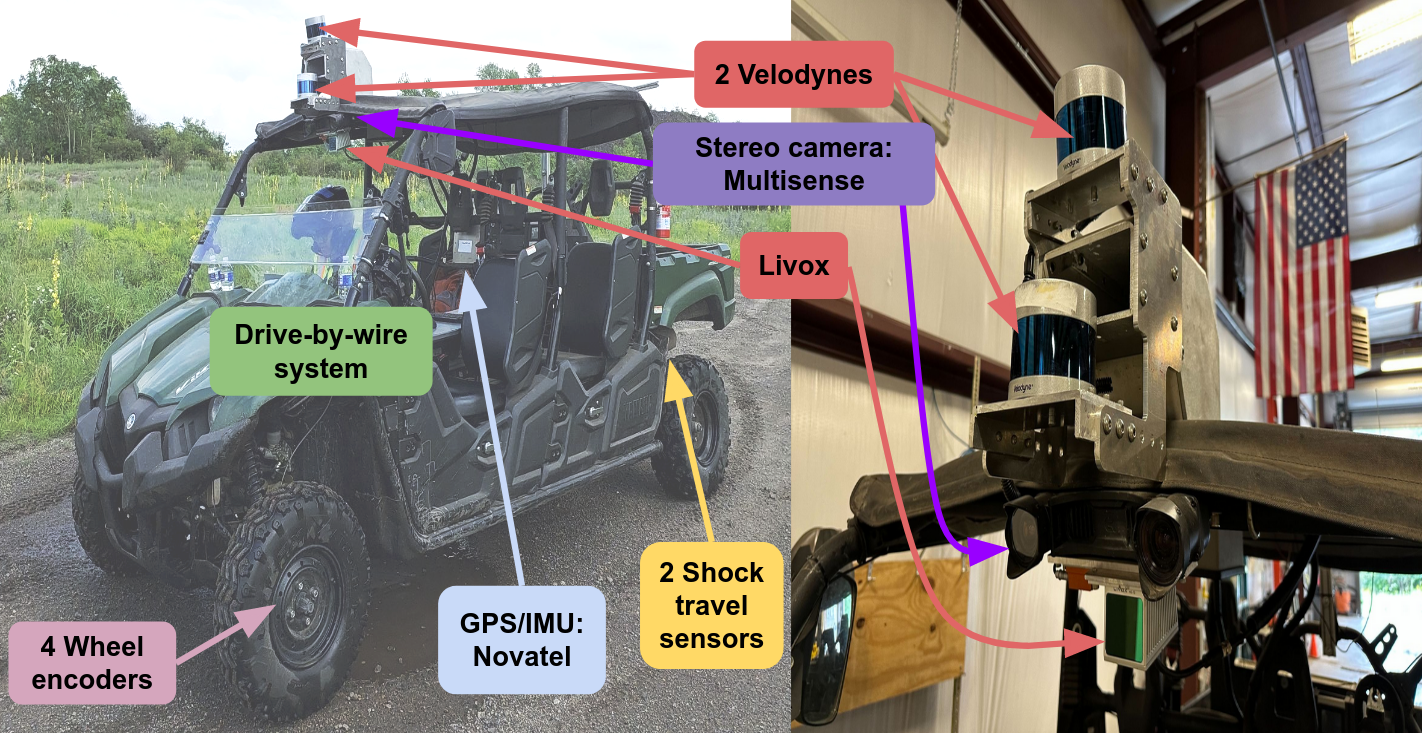}
    \caption{The ATV used for data collection (left); The primary sensor payload on the vehicle (right)}
    \label{fig:platform}
\end{figure}

\begin{figure}
    \centering
    \includegraphics[width=1.00\linewidth]{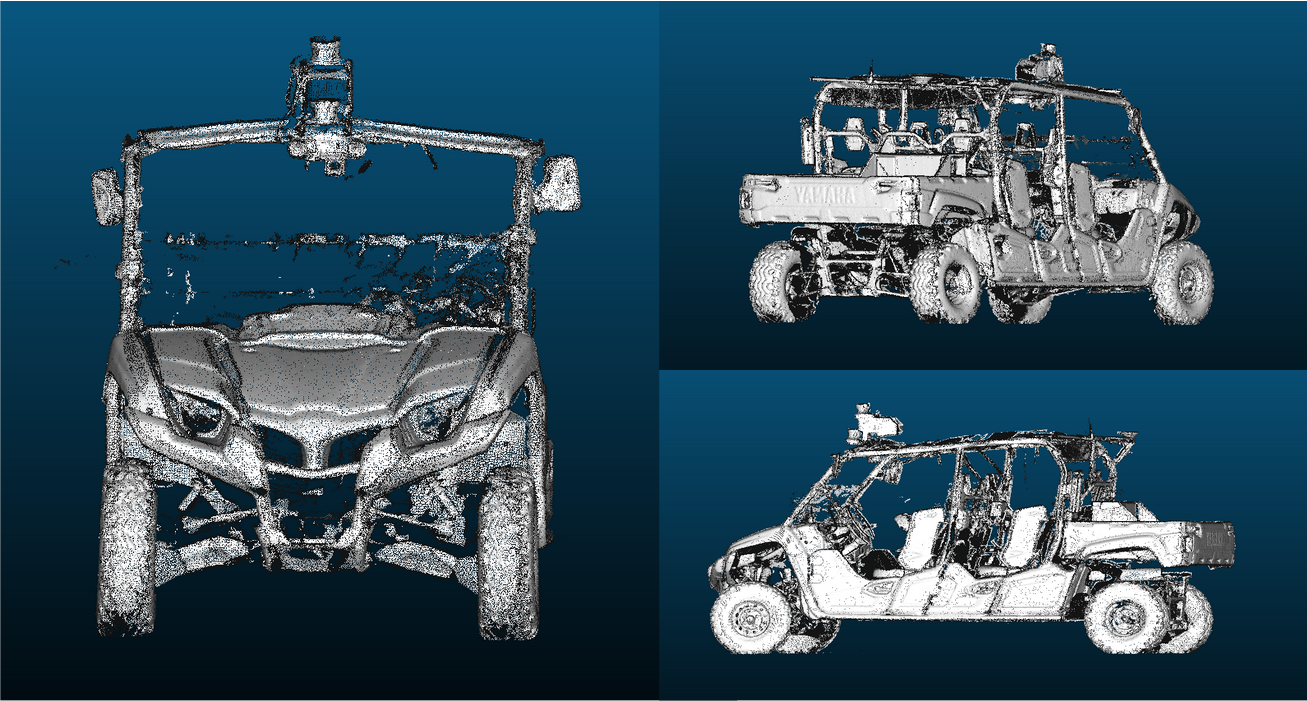}
    \caption{The 3D scan of the ATV. The rear half of the scan is cropped in the left picture to increase clarity of the front payload in the image.}
    \label{fig:scan}
\end{figure}

\section{The Dataset}
The original TartanDrive dataset was designed for the task of dynamics prediction. When designing TartanDrive 2.0 we aimed to support a broader set of tasks in perception, planning, and control. Thanks to more sensor modalities and better infrastructure, the new data that we have collected driving through diverse terrain and unique scenarios is more broadly applicable than before.

\subsection{The Platform}
We use the same Yamaha Viking All-Terrain Vehicle (ATV) as before (Fig. \ref{fig:platform}), first modified by Mai et al. \cite{Mai-2020}. We have modified it again in order to equip it with three LiDAR sensors. There are two Velodyne VLP-32 sensors mounted to the front of the roof of the vehicle, with one tilted downwards to increase coverage of the ground closer to the vehicle. There is also a Livox Mid-70, mounted under the MultiSense camera, which provides more information on objects directly in front of the vehicle. Using a Faro Focus Scanner, we generated a 3D pointcloud model of the ATV as shown in Fig. \ref{fig:scan} and provide it with the dataset so that users can take their own measurements (e.g. distance between the wheel axle and some arbitrary point on the vehicle, or distance between the GPS antenna and the ground) for their own specific experiments.

\subsection{Data Collection}
\begin{figure}
    \centering
    \includegraphics[scale=.3]{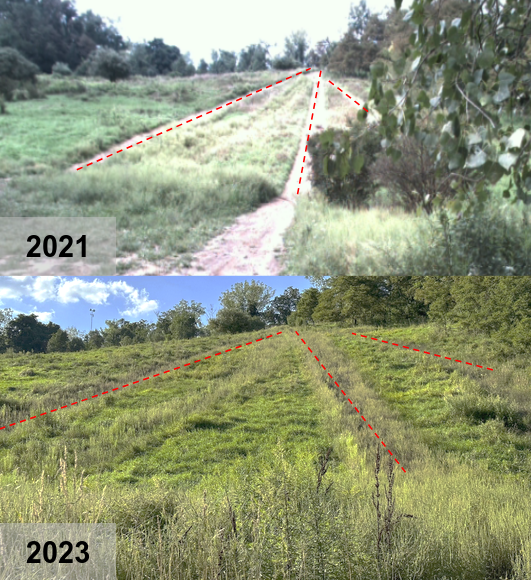}
    \caption{The data collection site has changed significantly since our last dataset released in 2021. For example, some older dirt paths (top, shown in red) are now covered in tall grass (bottom).}
    \vspace{-0.5cm}
    \label{fig:meadows_diff}
\end{figure}

The site for collecting data is the same location in western Pennsylvania as TartanDrive 1.0, consisting of terrain such as narrow paths, dense foliage, rocky terrain, dirt paths, and steep hills. It is worth noting that some areas of the site have changed significantly due to natural causes such as erosion or overgrowth as shown in Fig. \ref{fig:meadows_diff}. All the data included was acquired by a human tele-operating the vehicle. Each sequence of data is annotated with the following metadata:
\begin{itemize}
\item Driver ID and robot
\item Number of people in the vehicle
\item Date/time
\item Context (e.g. data collection)
\item Weather conditions (e.g. dry, damp, snow)
\item Lighting conditions (e.g. sunny, overcast, sunset)
\item Course ID or general location (from a pre-defined list)
\end{itemize}
as well as any other information relevant to that specific run. We emphasize that recording this metadata makes the dataset significantly easier to use, especially as the dataset becomes larger. At the end of the collection, a post-processor records information such as top speed, average speed, duration, and sensors present. During data collection, a co-pilot takes time-stamped annotations of relevant or unique events such as sensor failure or something uncommon such as a deer spotting in front of the vehicle. Additionally, in some runs the co-pilot periodically annotates a weak driving score ranging from 1-5 where 5 signifies ideal driving. Recording this metadata provides more structure in the data which significantly improves the utility of the dataset when used in combination with our post-processing pipelines.

\begin{figure}
    \centering
    \includegraphics[width=.80\linewidth]{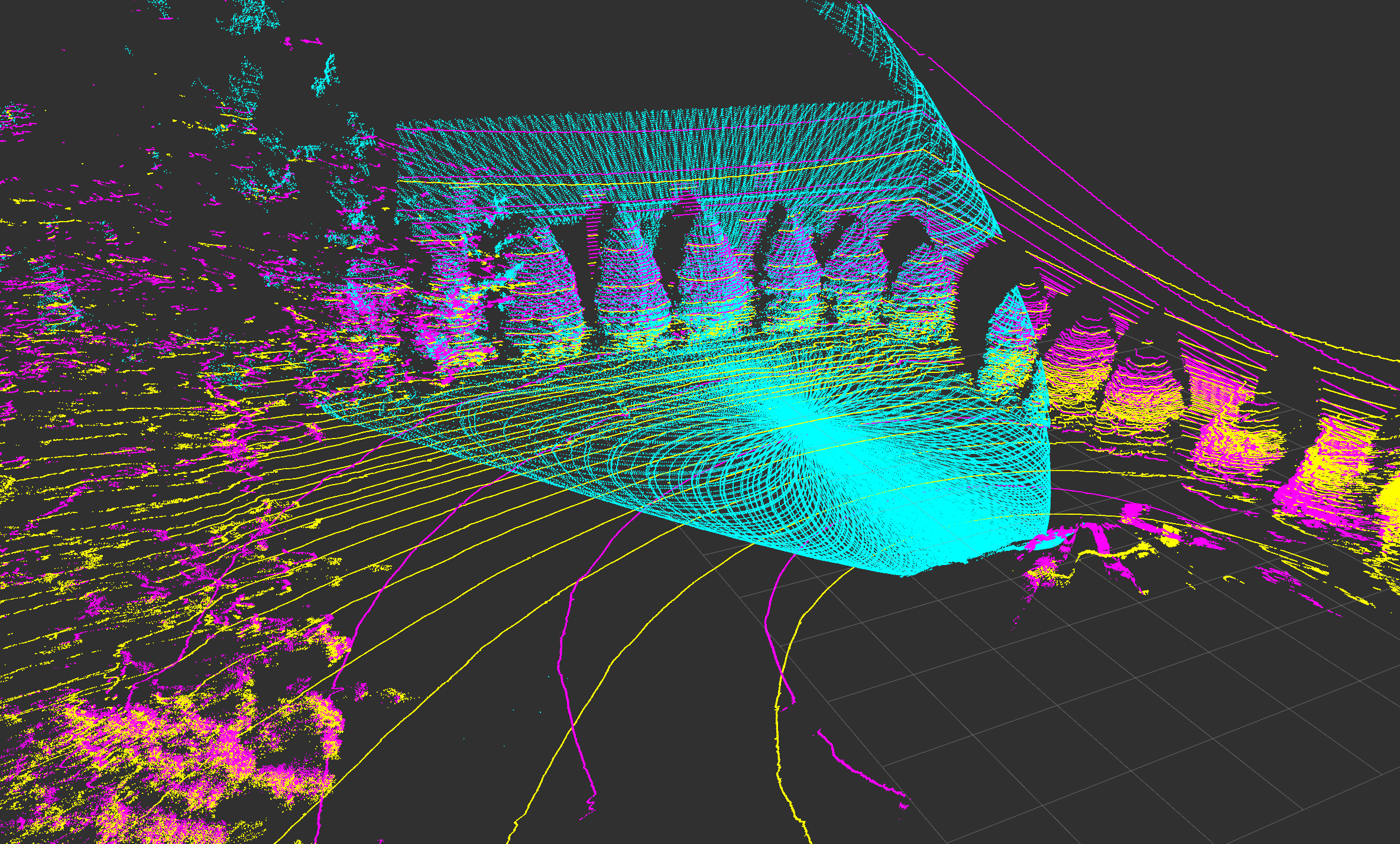}
    \caption{Example coverage provided by our LiDAR sensors. Purple and yellow points come from the two Velodynes, and cyan points from the Livox.}
    \label{fig:lidar_coverage}
\end{figure}
\vspace{-0.2cm}

\begin{figure*}
    \centering
    \includegraphics[width=.8\textwidth, height=185pt]{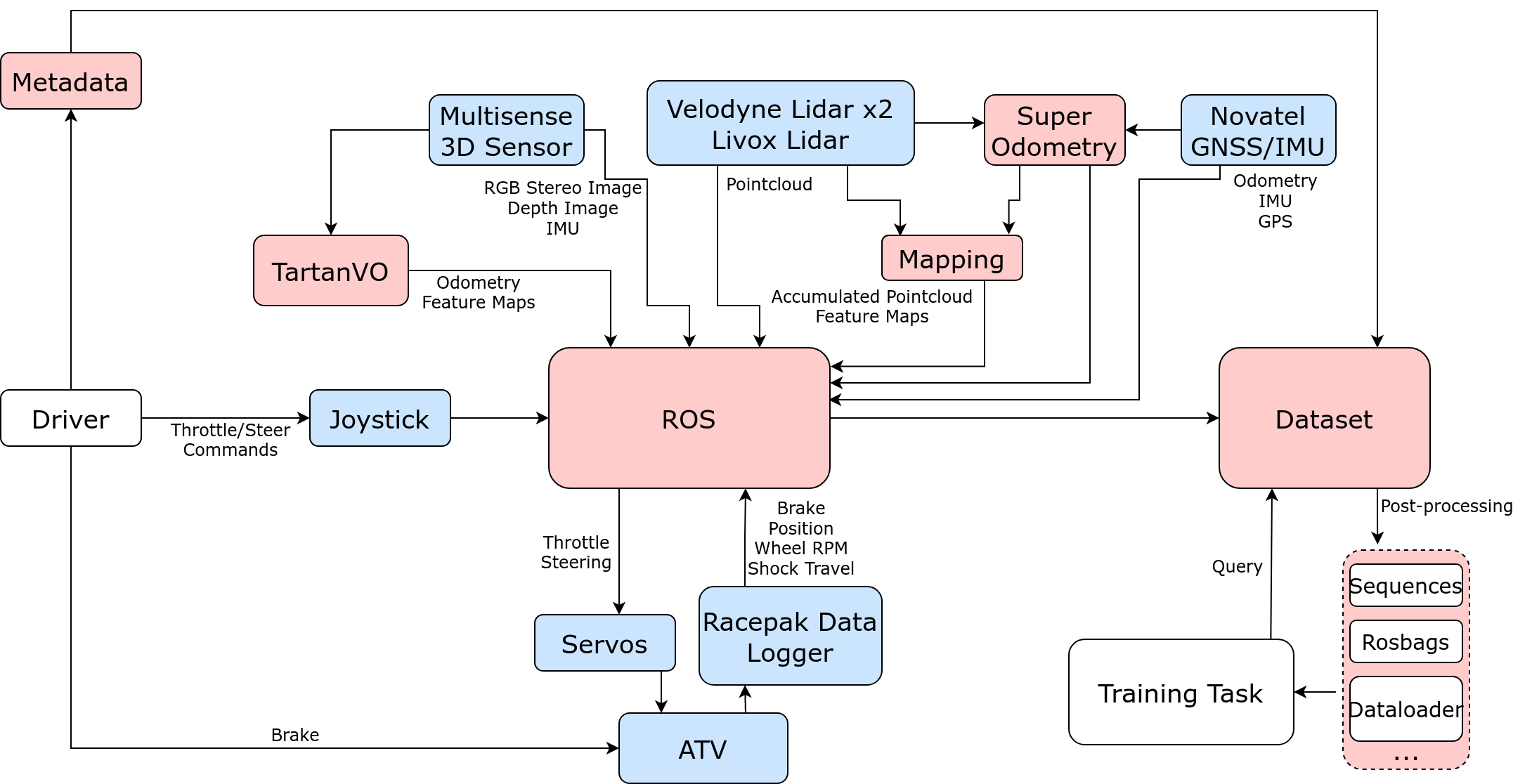}
    \caption{The high-level flow of information in our data collection process, with software in red and hardware in blue.}
    \label{fig:flowchart}
\end{figure*}

\subsection{Raw Data}
Most of the sensors on our platform have already been detailed \cite{9811648,sivaprakasam2023tartandrive}, but all raw data is briefly summarized below:
\subsubsection{Pointclouds}
We record incoming pointclouds from two Velodyne VLP-32 LiDAR sensors and a Livox Mid-70. We also provide extrinsics so that they can be merged into a single pointcloud. An example of the coverage they provide is shown in Fig. \ref{fig:lidar_coverage}. All LiDARs are configured to run at 10Hz.

\subsubsection{Images}
A Carnegie Robotics MultiSense S21 is used to provide stereo images, specifically greyscale images from both cameras as well as RGB images from the left camera at 10Hz each.

\subsubsection{IMU and Pose}
A NovAtel PROPAK-V3-RT2i GNSS provides IMU data at 100Hz. This is fused with incoming GPS data to provide a pose estimate at 50Hz. The MultiSense also provides IMU data at 400Hz.

\subsubsection{Teleoperation}
The driver uses a joystick controller to send steering commands, which are then recorded alongside the values of the current steering angle (actuation on steering commands is not instantaneous). A Racepak G2X Pro Data Logger is used to provide the positions of the acceleration and brake pedals.

\subsubsection{Proprioceptive Information}
The Racepak is also used to record RPMs for each wheel and suspension shock travel for the rear two wheels. Shock travel data for the front two wheels is omitted due to frequent damage to the sensors during aggressive driving maneuvers, but will be included in future data releases.

\subsection{Post-Processed Data}


In order to increase utility and ease-of-use of the dataset, we provide the raw data and the outputs from some existing modules that have been integrated into our software stack (Fig. \ref{fig:flowchart}) and commonly use as inputs to our own algorithms:

\subsubsection{TartanVO}
We run TartanVO \cite{tartanvo2020corl} on the platform, which takes in the stereo images from the MultiSense as input. It outputs an odometry estimate, a predicted pointcloud, and top-down height and RGB maps, all at 10Hz.


\subsubsection{Roughness Cost}
We produce a roughness cost that describes the bumpiness of terrain as we drive over it, derived from a sliding window of Z-axis linear acceleration values from the IMU as described in Guaman Castro et al. \cite{castro2023does}.

\subsubsection{Odometry and Registered Pointcloud}
In addition to the odometry provided by the Novatel system, we also provide an estimated output by Super Odometry \cite{superodometry}. The high accuracy and frequency of this output allows for cleaner results in tasks such as pointcloud registration which is in turn important for other downstream tasks.

\subsubsection{Local LiDAR Maps}
\begin{figure}
    \centering
    \includegraphics[scale = .13]{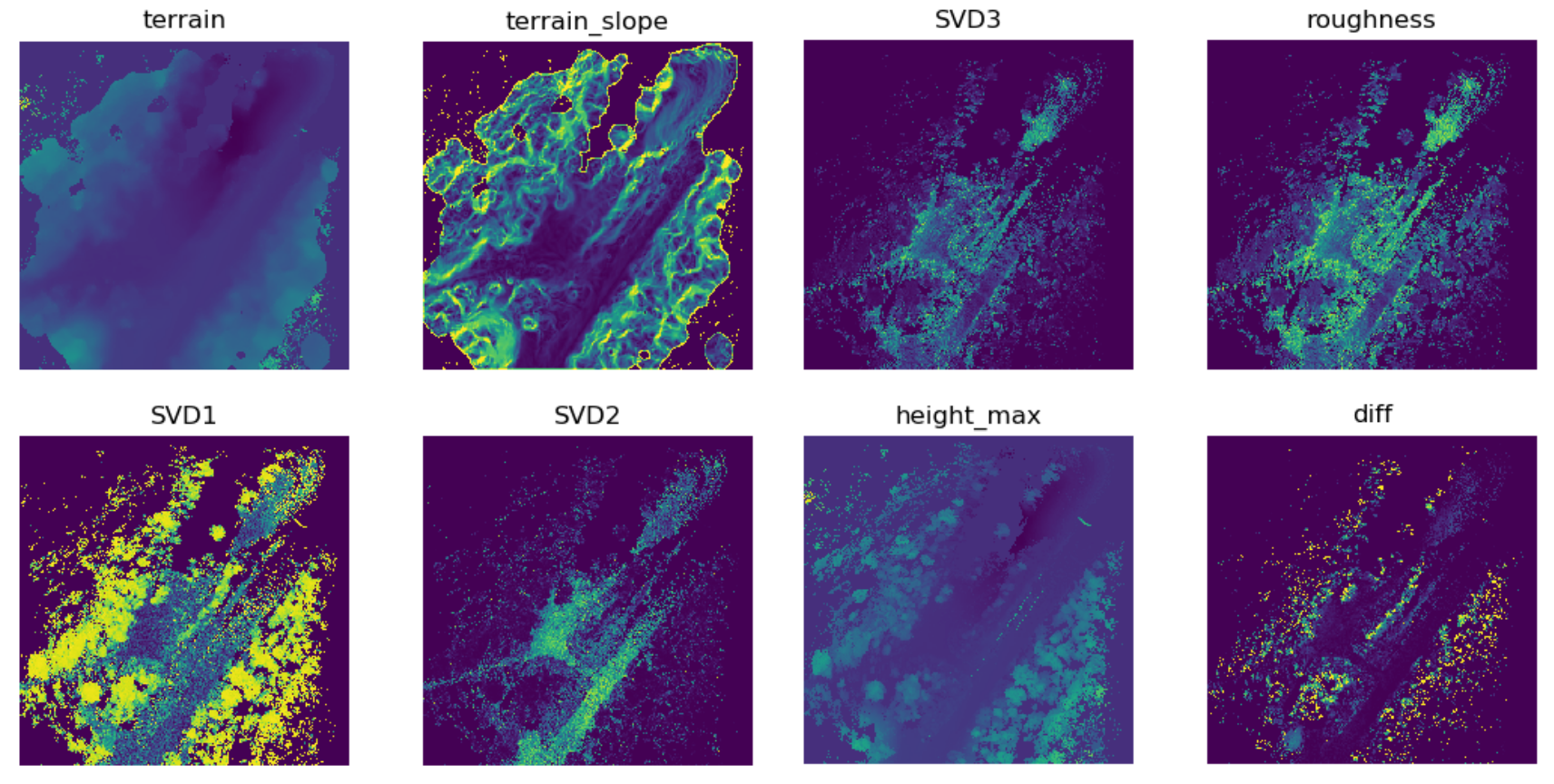}
    \caption{Using the registered pointcloud, we provide local maps that contain various geometric features.}
    \vspace{-0.5cm}
    \label{fig:feat_maps}
\end{figure}

Using the registered pointcloud provided by Super Odometry, we generate a birds-eye-view feature map 200x200m wide at .5m resolution. This map provides geometric information about the environment (Fig. \ref{fig:feat_maps}), with features consisting of the following:
\begin{itemize}
\item Min/Max/Mean Height of Points
\item Roughness
\item SVD Features
\item Estimated Ground Height
\item Estimated Ground Slope (X, Y, Magnitude)
\end{itemize}

\begin{figure*}
    \centering
    \includegraphics[width=.9\linewidth,height=5.8cm]{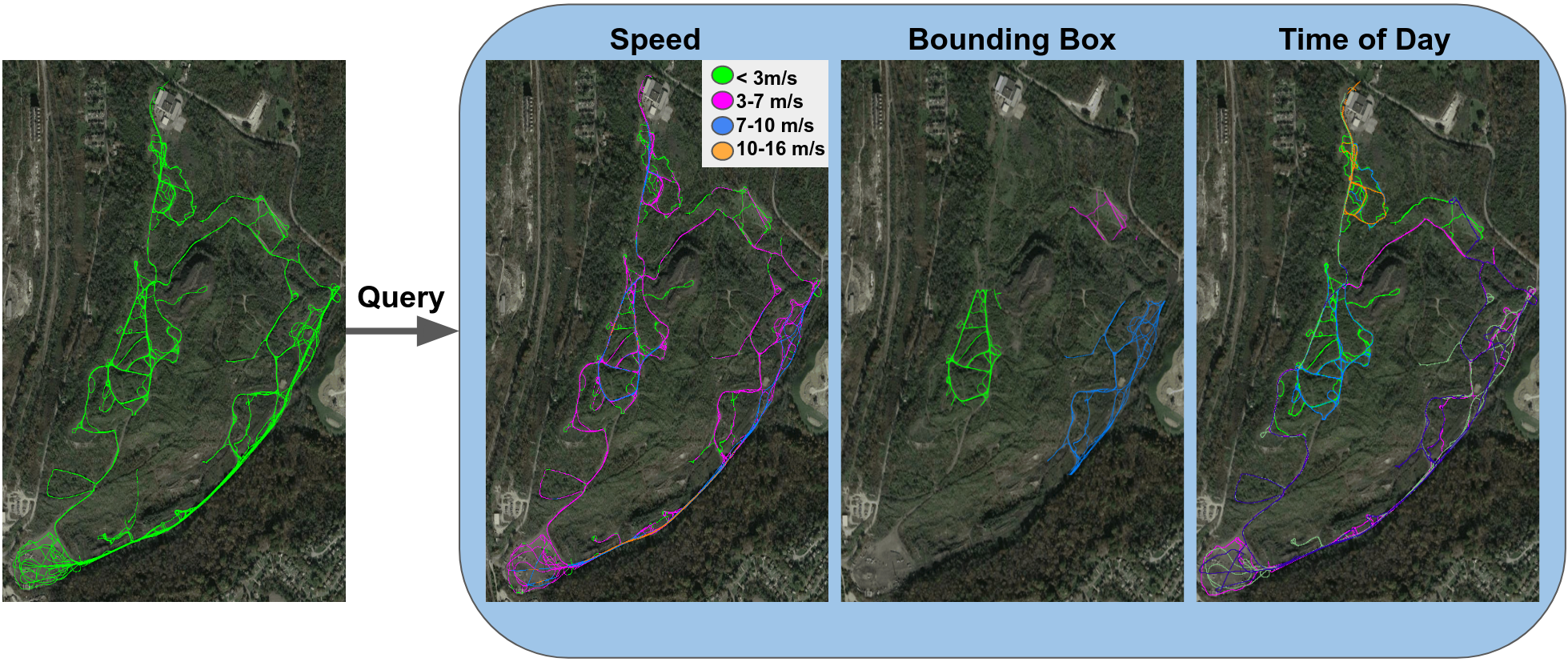}
    \caption{Our dataset covers over 255 acres collected over seven hours. Using the accompanying metadata, the it can easily be split into groups by aspects such as speed, GPS bounding box, and time of day.}
    \label{fig:metadata_splitting}
\end{figure*}

\subsection{Data Pipelines}
\subsubsection{Formatting}
We provide the data in two main formats, the first one being the rosbags that they were originally recorded as. This allows users to test how their algorithms might perform in different environments in real-time. The second format is as a set of sequences similar to the KITTI format \cite{Geiger2013IJRR}. For each bag, a folder is created and a subfolder for each modality is initialized. The samples across all modalities are then timesynced and then placed in their respective subfolders (e.g. 'pointcloud\_1/0000.bin' and 'image/0000.jpg' are associated with each other).

\subsubsection{Reconfiguration}

We have post-processed our data in a way that makes sense for our own algorithms, but our ATV platform is somewhat unique. Across different robots, the optimal parameters for elements such as sample frequency, map size, and map resolution vary. For example, a smaller robot with better agility might require a map with finer than .5m resolution. In order to make our data more directly applicable to other platforms, we provide scripts that allow users to regenerate the dataset with different parameters as they see fit.

\subsubsection{Utilities}

As previously mentioned, we collect metadata and some annotations during data collection. We provide scripts to take advantage of this metadata in order to filter the data by various elements and therefore increase the utility of the dataset for different learning tasks. For example, given the whole dataset, a user can easily create subsets grouped by components such as speed, GPS bounding box (we provide a GUI for easily generating bounding boxes), lighting conditions, or driver (Fig. \ref{fig:metadata_splitting}). This is especially useful for testing model generalizability by training on one location/condition, and testing on another. 

\subsubsection{Common Framework}
The infrastructure we have developed has streamlined our data collection procedures to make it easy to train models on processed datasets as well as test our algorithms in real-time. By continuing to follow the same procedures, the data we collect in the future can easily be merged into our existing datasets. Likewise, other researchers can use our tools when collecting their own off-road driving data which could eventually lead to a larger multi-site, and multi-vehicle dataset. 

\begin{figure} [t]
    \centering
    \includegraphics[width=.70\linewidth]{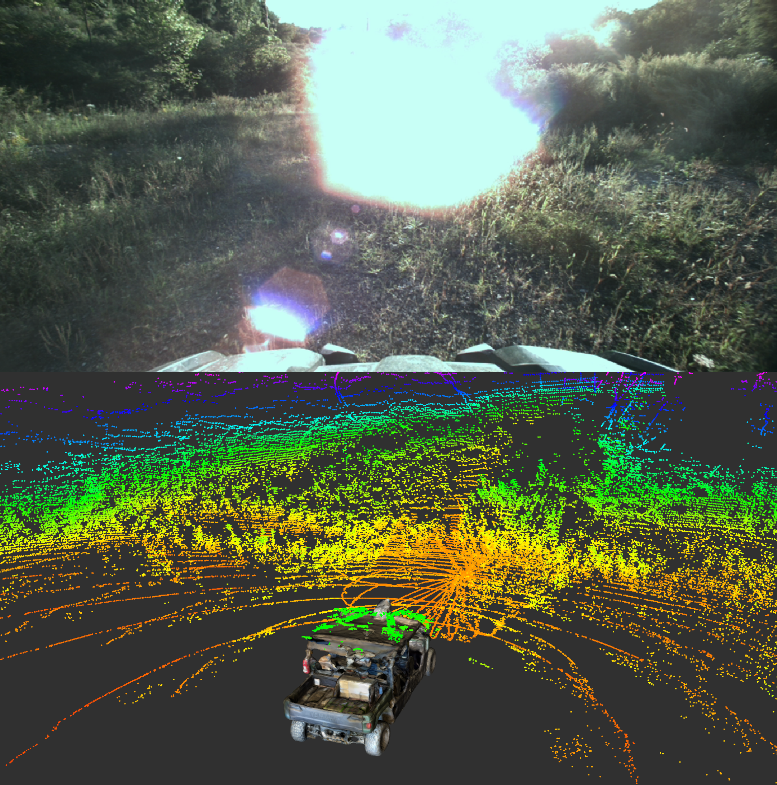}
    \caption{In cases where conditions like poor lighting affect camera performance, LiDARs can allow a robot to still understand its surroundings.}
     \vspace{-0.5cm}
    \label{fig:lens_flare}
\end{figure}

\begin{figure} [t]
    \centering
    \includegraphics[width=.80\linewidth,height=6cm]{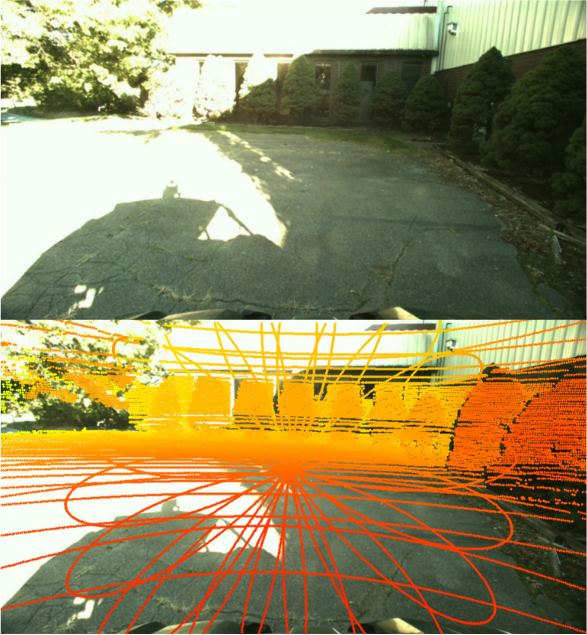}
    \caption{An example of the overlap between a single scan from all 3 LiDARS and our camera. Points are colored by forward distance from the vehicle.}
     \vspace{-0.5cm}
    \label{fig:camera_lidar}
\end{figure}

\section{Impact on Current and Future Research}
While we do not provide any explicit supervision labels (e.g. segmentation masks, classification labels) outside of annotations, the several modalities and actions that we provide facilitate a number of self-supervised learning tasks. Our original TartanDrive has already benefited off-road research immensely, and we believe the categories listed in Table \ref{tab:table} are only a subset of the research topics TartanDrive 2.0 can bolster.

\begin{table*}[t]
    \centering
    \vspace*{0.1in}
    \begin{tabular}{c||c|c|c|c|c|c|c|c|c|c|c}
         Dataset & State & Action & Image & Pointcloud & Heightmap & RGBmap & IMU & Wheel RPM & Shocks & Metadata & Labels\\
         \hline
         RUGD \cite{RUGD2019IROS} & No & No & Yes & No & No & No & No & No & No & No & Yes\\
         Rellis 3D \cite{jiang2020rellis3d}  & Yes & Yes & Yes & Yes & No & No & Yes & No & No & No  & Yes
         \\
         Wild-Places \cite{knights2023wildplaces}  & Yes & No & No & Yes & No & No & No & No & No & No  & Yes
         \\
         Montmorency \cite{tremblay2019automatic} & Yes & Yes & Yes & Yes & Yes & No & Yes & No & No & No  & Yes\\
         Verti-Wheelers \cite{verti-wheelers} & Yes & Yes & Yes & No & No & No & Yes & Yes & No & No & No\\
         TartanDrive 1.0 \cite{9811648} & Yes & Yes & Yes & No & Yes & Yes & Yes & Yes & Yes & No & No\\
         TartanDrive 2.0 (Ours) & Yes & Yes & Yes & Yes & Yes & Yes & Yes & Yes & Yes & Yes & No
    \end{tabular}
    \caption{Overview and comparison of various off-road driving datasets}
    \label{tab:table}
\end{table*}
\begin{figure}
    \centering
    \includegraphics[width=1.0\linewidth,height=7cm]{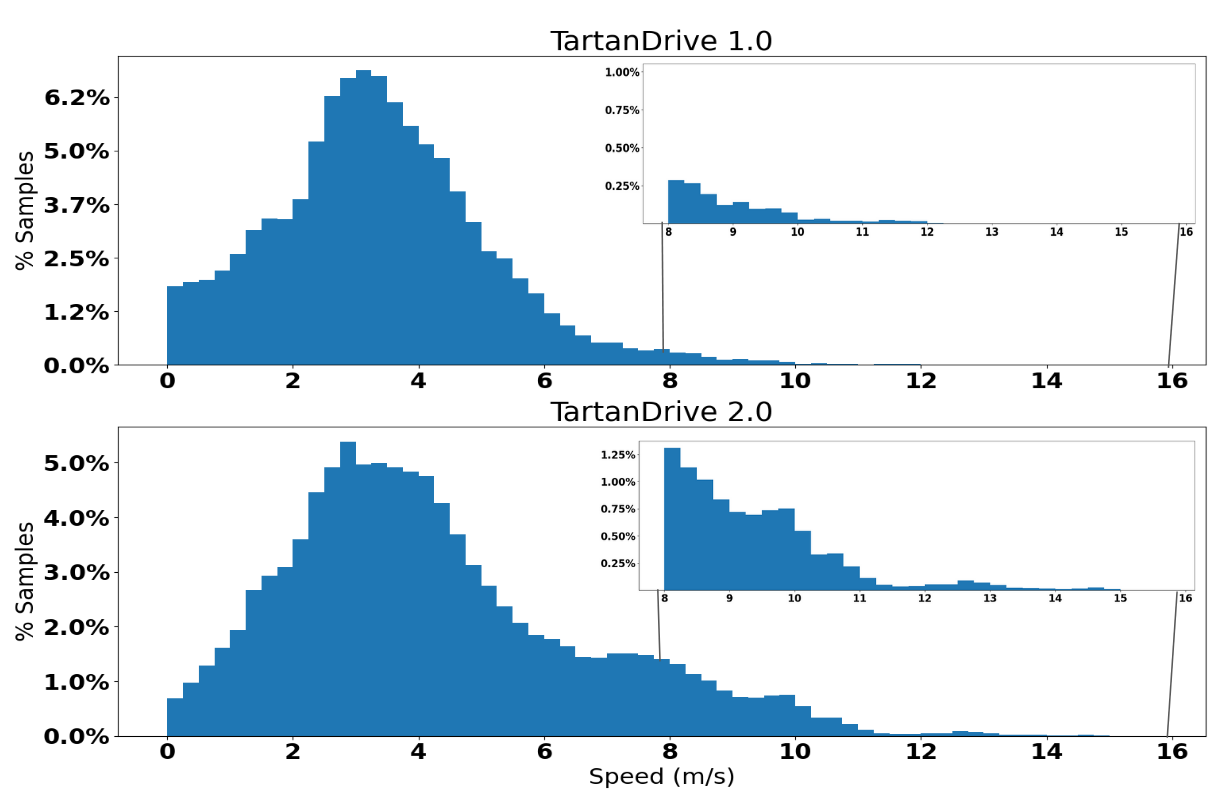}
    \caption{In TartanDrive 2.0 we collect a higher concentration of data at higher speeds than in TartanDrive 1.0, reaching up to 15m/s.}
    \vspace{-0.5cm}
    \label{fig:speed_distribution}
\end{figure}

\subsection{Perception - Cross-Modal Supervision:}
Our data was used by Guaman et al. to visually predict a bumpiness cost supervised by an IMU-derived metric \cite{castro2023does}. With the addition of LiDAR in our dataset, we can advance algorithms by providing information in scenarios where cameras fail (Fig. \ref{fig:lens_flare}) and learning from a much more accurate source of odometry and depth (Fig. \ref{fig:camera_lidar}). For example, Chen et al. accumulates near-range LiDAR measurements to learn long-range visual traversability model \cite{chen2023learningonthedrive}. Meng et al. train a model that uses images to predict feature maps supervised by lidar inputs \cite{meng2023terrainnet}. However, these works rely on off-road driving datasets that are not publicly available. By releasing data in a similar domain with the same modalities, we lower the barrier to entry for expanding on this field of research, and allow a common point of comparison.
\subsection{Perception - Map Completion:}
Prediction of occluded and sparsely sensed areas enables faster safe navigation in occluded area \cite{meng2023terrainnet,schmid2022selfsupervised}. Our generated accumulated maps are being used as a supervisory signal in our ongoing research in map completion and can also serve as a common benchmark for other approaches.
\subsection{Perception - Ground Height Estimation:}
An understanding of the support ground surface in off-road terrain is essential for successful navigation. We enable methods of learning this feature by providing multiple modalities as inputs and a 3D model of the car that can be used alongside odometry estimates to provide supervision of the true ground surface.
\subsection{Planning - Learning from Demonstration:}
The actions derived from human teleoperation coupled with sensor inputs can be used to learn models that can reason about what areas are more traversable than others. Triest et al. treats teleoperated driving as expert demonstrations and use inverse reinforcement learning to birds-eye-view costmaps from the LiDAR feature maps \cite{triest2023learning}.
\subsection{Controls - Aggressive Maneuver Driving:}
The unstructured nature of complex terrain makes it difficult to drive aggressively at high speeds without losing control. Our new dataset contains a higher proportion of speeds beyond 7m/s than before. Some of our ongoing research on learning vehicle dynamics models is supported by this data.

\section{Conclusions and Future Work}

We have presented TartanDrive 2.0, which provides significant improvements over the original TartanDrive. We emphasize how the inclusion of more modalities can benefit research in self-supervised learning, and how more data and better infrastructure allow it to scale up to the needs of large models. The tools that we release with the dataset have streamlined our data collection, facilitating our plans of collecting more data on our existing site as the seasons change and then releasing that data over time.

Despite the major improvements in this generation of our dataset, there remains a number of directions to pursue in order to further increase its quality and utility. For example, while LiDAR sensors are useful in much of modern robotics research, there has also been an increasing amount of research in using passive sensors for perception \cite{srivastav2023radars,Brenner_2023}. To that end it would be helpful to add sensors like thermal cameras in the near future.

We also plan to include audio and language in our data. There is an increasing amount of research in foundation models and vision-language models \cite{openai2023gpt4, clip}, some of which have already demonstrated improvement in some robotics applcations \cite{rt12022arxiv, vemprala2023chatgpt, shridhar2022cliport, shah2023lm}. We hypothesize that learning the relationship between driver dialogue and their actions and/or the resulting environmental interactions might be supplemental in training large neural networks. Apart from recording dialogue, we also plan to record audio of the interactions between tires and the ground as we drive through different terrains, as Zurn et al. have shown that this information can be used to learn an understanding of complex terrain \cite{zurn}.

Additionally, our test site is very large and complex, but we have found that our own algorithms are close to enabling the ATV to successfully traverse much of the terrain. This brings up the issue of overfitting, and whether or not our algorithms would be able to reach the same level of performance in unseen, challenging terrain. By releasing our tools alongside the dataset we hope that others will contribute data from other locations in order to collaboratively form an even larger dataset. However, regardless of external contributions, we plan to collect data in other off-road sites with different types of challenging terrain that will allow ourselves and others to test generalizability and online adaptation of our algorithms. 

\newpage


{
\bibliographystyle{IEEEtran}
\bibliography{refs}
}

\end{document}


\appendix

\subsection{Detailed Algorithms}

We present in detail the algorithms described in the main paper. Algorithm \ref{algo:mppi_state_visitations} describes the algorithm for extracting state visitations from an MPPI solution. Algorithm \ref{algo:medirl_train} describes the training step for our algorithm. Algorithm \ref{algo:medirl_test} describes the risk-aware costmap generation process.

\begin{algorithm}
\DontPrintSemicolon
\caption{Computation of State Visitation Frequencies (SVF) from Weighted Trajectories}
\label{algo:mppi_state_visitations}
\KwIn{Initial State $x_0$, Weights $\boldsymbol{\eta}$, Trajectories $\boldsymbol{\tau}$, Map resolution $M_{res}$}
\KwOut{State Visitation Frequencies $D$}

\For{$\tau_n, \eta_n \in \boldsymbol{\tau}, \boldsymbol{\eta}$}{
    \For{$x_t \in \tau_n$} {
        $i, j \leftarrow \left \lfloor {p(x_t) / M_{res}} \right \rfloor$ \;
        $D[i, j] \leftarrow D[i, j] + \eta_n$ \;
    }
}

$D  \leftarrow \frac{D}{\sum_{i, j} D[i, j]}$ \;

return $D$ \;

\end{algorithm}

\begin{algorithm}
\DontPrintSemicolon
\caption{Training Step for Fast MEDIRL}
\label{algo:medirl_train}
\KwIn{\\
        \quad Dataset $\mathcal{D}$ of expert trajectories $\boldsymbol{\tau_E}$, \\
        \quad Map features $\mathbf{M}$ , \\
        \quad Goal weight $\kappa$, \\
        \quad FCN Ensemble $F_\theta(M) : \mathcal{R}^{W \times H \times D} \rightarrow \mathcal{R}^{B \times W \times H}$, \\
        \quad MPPI $MPPI(x_s, x_g, C, \lambda): \mathcal{X} \rightarrow (\eta_n, \tau_n) \times N$}

\While{not converged}
{
    $\tau^E, M \sim \mathcal{D}$ \hfill $\triangleleft$ Sample from dataset \;
    $x_0 = \tau^E_0, x_g = \tau^E_{t-1}$ \hfill $\triangleleft$ set start/goal \;
    $f_\theta \sim \boldsymbol{F_\theta}$ \hfill $\triangleleft$ sample FCN from ensemble \;
    $C = f_\theta(M_0)$ \hfill $\triangleleft$ Compute costmap from FCN \;
    $\boldsymbol{\tau^L}, \boldsymbol{\eta^L} = MPPI(x_0, x_g, C, \kappa)$ \;
    $D^E = SVF(\tau_E)$ \hfill $\triangleleft$ Compute state visitations \;
    $D^L = SVF(\tau^L, \eta^L)$ \quad via Algorithm \ref{algo:mppi_state_visitations}\;
    $\nabla_z J = D^E - D^L$ \hfill $\triangleleft$ Gradient via \cite{wulfmeier2015deep} \;
    backprop($\nabla_z J, f_\theta$) \hfill $\triangleleft$ Update FCN grads\;
    $\theta \leftarrow \text{Adam}(\theta - \nabla_\theta J)$ \hfill $\triangleleft$ Update via \cite{kingma2014adam}\;
}
\end{algorithm}

\begin{algorithm}
\DontPrintSemicolon
\caption{Inference Step for Fast MEDIRL}
\label{algo:medirl_test}
\KwIn{\\
        \quad Map features $M$, \\
        \quad FCN Ensemble $F_\theta(M) : \mathcal{R}^{W \times H \times D} \rightarrow \mathcal{R}^{B \times W \times H}$, \\
        \quad Risk level $\nu$}

$C_\nu = \boldsymbol{0}^{m \times n}$ \hfill $\triangleleft$ Initialize empty costmap\;
$\boldsymbol{C} = \boldsymbol{F_\theta}(M)$ \hfill $\triangleleft$ \hfill $\triangleleft$ Create costmap for each FCN\;
\For{$i,j \in ([0 \hdots m] \times [0 \hdots n])$}
{
    $\boldsymbol{c} = \boldsymbol{C}[:, i, j]$ \hfill $\triangleleft$ Get each FCN's cell output \;
    $C_\nu[i, j] = CVaR_\nu(\boldsymbol{c})$ \hfill $\triangleleft$ Compute CVaR \;
}

return $C_\nu$

\end{algorithm}

\subsection{Lidar Mapping Algorithm}

Our lidar mapping algorithm is presented in detail in Algorithm \ref{algo:lidar_mapping}. 

\begin{algorithm*}
\DontPrintSemicolon
\caption{Lidar Mapping Algorithm}
\label{algo:lidar_mapping}
\KwIn{Buffer of registered pointclouds $P_b$, pointcloud skip $k_p$, Map origin $(o_x, o_y)$, Map size $(l_x, l_y)$, Map resolution $r$, overhang limit $k_{overhang}$}
\KwOut{Terrain feature tensor $X$}

$n_x = \lfloor \frac{l_x}{r} \rfloor$ \;
$n_y = \lfloor \frac{l_y}{r} \rfloor$ \;
$X = \mathbf{0} ^ {n_x \times n_y \times 12}$ \hfill $\triangleleft$ Initialize map tensor \;
$P = \sum_{i=0}^{|P_b| / k_p} P_{i * k_p}$  \hfill $\triangleleft$ Aggregate pointclouds from buffer\;

\For{$i=0 \hdots n_x$}{
    \For{$j=0 \hdots n_y$}{
        $P_m = \{p, \forall p \in P | (\frac{P_x - o_x}{r} = i) \land (\frac{P_y - o_y}{r} = j)\}$  \hfill $\triangleleft$ Get all points in a given column \;
        
        $X[i, j, 0] = min_{p \in P_m} [p_z]$ \hfill $\triangleleft$ Get min height \;
        $X[i, j, 1] = max_{p \in P_m} [p_z]$  \hfill $\triangleleft$ Get max height \;
    }
}

$X[:, :, 4] = G * inflate(X[:, :, 0])$ \hfill $\triangleleft$ Generate terrain estimate by inflating and low-pass filtering min height \;
$X[:, :, 5] = S_x * |X[:, :, 2]| + S_y * |X[:, :, 2]|$ \hfill $\triangleleft$ Get terrain slope via derivative filter\;

\For{$i=0 \hdots n_x$}{
    \For{$j=0 \hdots n_y$}{
        $P_m = \{p, \forall p \in P | (\frac{p_x - o_x}{r} = i) \land (\frac{p_y - o_y}{r} = j) \land (p_z < X[i, j, 2] + k_{overhang})\}$ \hfill $\triangleleft$ Filter overhanging points \;
        
        $X[i, j, 2] = \max_z [p_z \forall p \in P_m]$ \hfill $\triangleleft$ Get the max height of the cell, saturating at the overhang limit\;

        $X[i, j, 3] = \frac{1}{|P_m|} \sum_{p \in P_m} [p_z]$ \hfill $\triangleleft$ Get the mean height of the cell\;

        $X[i, j, 6] = X[i, j, 2] - X[i, j, 4]$ \hfill $\triangleleft$ Get the height of the cell relative to terrain \$

        $\lambda_1, \lambda_2, \lambda_3 = SVD(P_m)$ \hfill $\triangleleft$ Get the SVD decomposition of the cell points \;
        $X[i, j, 7] = \frac{\lambda_1 - \lambda_2}{\lambda_1}$ \hfill $\triangleleft$ Get SVD1 \;
        $X[i, j, 8] = \frac{\lambda_2 - \lambda_3}{\lambda_1}$ \hfill $\triangleleft$ Get SVD2 \;
        $X[i, j, 9] = \frac{\lambda_3}{\lambda_1}$ \hfill $\triangleleft$ Get SVD3 \;
        $X[i, j, 10] = \frac{\lambda_3}{\lambda_1 + \lambda_2 + \lambda_3}$ \hfill $\triangleleft$ Get roughness \;
        $X[i, j, 11] = \mathbf{1}[|P_m| = 0]$ \hfill $\triangleleft$ Get unknown \;
    }
}

return $X$ \;
\end{algorithm*}

\subsection{Navigation Course Overview}

A more detailed description of the course used for the navigation experiments and representative terrains is presented in Figure \ref{fig:nav_course}.

\begin{figure}
    \centering
    \includegraphics[width=\linewidth]{media/IRL media/nav_course.png}
    \caption{An overview of our navigation course. In total, the course was roughly 1.6km, and included several challenging scenarios such as going over slopes and through tall grass. The course began and ended at the green diamond, and followed the red arrows. The path is colored according to elevation change.}
    \label{fig:nav_course}
\end{figure}

\subsection{Comparison of CVaR Plots}

Presented below are additional qualitative visualizations of the costmaps learned from linear features and resnets. Visualizations are produced from representative examples of terrain from the test set in Figures \ref{fig:resnet_cvar_qual} and \ref{fig:linear_cvar_qual}. Note that \textit{each} subplot has its own color normalization. That is, changes in color denote changes in cost \textit{relative} to other costs in that particular subplot. This was done to mitigate higher CVaR potentially producing a constant bias and shift. Also note that the vehicle is located in the center of each BEV plot, and the orientation of the vehicle in the map is notated via the red arrow.

\begin{figure*}
     \centering
     \begin{subfigure}[t]{\linewidth}
         \centering
         \includegraphics[scale=0.4]{media/IRL media/resnet_00609.png}
         \caption{Corridor Scenario. The relative cost of obstacles and trail are consistent for all $\text{CVaR}_\nu$, while the cost of unknown space increases.}
     \end{subfigure}
     \hfill
     \begin{subfigure}[t]{\linewidth}
         \centering
         \includegraphics[scale=0.4]{media/IRL media/resnet_01320.png}
         \caption{Tall Grass Scenario. The relative cost of grass increases with $\text{CVaR}_\nu$.}
     \end{subfigure}
     \hfill
     \begin{subfigure}[t]{\linewidth}
         \centering
         \includegraphics[scale=0.4]{media/IRL media/resnet_01649.png}
         \caption{Slope Scenario. The relative cost of the slope in the FPV increases with $\text{CVaR}_\nu$.}
     \end{subfigure}
     \hfill
     \caption{Resnet Results on representative terrains in the test set. Note that in uncertain regions such as grass and slopes, the resnet-based costmap adjusts costs based on CVaR.}
     \label{fig:resnet_cvar_qual}
\end{figure*}

\begin{figure*}
     \centering
     \begin{subfigure}[t]{\linewidth}
         \centering
         \includegraphics[scale=0.4]{media/IRL media/linear_00609.png}
         \caption{Corridor Scenario}
     \end{subfigure}
     \hfill
     \begin{subfigure}[t]{\linewidth}
         \centering
         \includegraphics[scale=0.4]{media/IRL media/linear_01320.png}
         \caption{Tall Grass}
     \end{subfigure}
     \hfill
     \begin{subfigure}[t]{\linewidth}
         \centering
         \includegraphics[scale=0.4]{media/IRL media/linear_01649.png}
         \caption{Slope Scenario}
     \end{subfigure}
     \hfill
     \caption{Linear Results on representative terrains in the test set. Note that the costmaps are more or less unchanged with respect to CVaR.}
     \label{fig:linear_cvar_qual}
\end{figure*}

\subsection{KBM Dynamics}
The full form of our KBM dynamics are presented in Equation \ref{eq:kbm}, with hyperparameters in Table \ref{tab:model_params}. Essentially, we use a standard KBM with velocity and steering setpoints as controls.

\begin{equation}
    \label{eq:kbm}
    \mathbf{X} = 
    \begin{bmatrix}
        x \\
        y \\
        \theta \\
        v \\
        \delta \\
    \end{bmatrix},
    \mathbf{U} =
    \begin{bmatrix}
        v_{target} \\
        \delta_{target}
    \end{bmatrix},
    \mathbf{\dot{X}} = 
    \begin{bmatrix}
        v cos(\theta) \\
        v sin(\theta) \\
        v \frac{tan(\delta)}{L} \\
        K_v (v_{target} - v) \\
        K_\delta (\delta_{target} - \delta) \\
    \end{bmatrix}
\end{equation}

\subsection{Hyperparameters}
The hyperparameters used in our experiment are provided in Tables \ref{tab:model_params} and \ref{tab:mppi_params}.

\begin{table}[]
    \centering
    \begin{tabular}{c|c}
        Parameter & Value \\
        \hline
         $L$ & $3.0m$ \\
         $K_v$ & $1.0$ \\
         $K_\delta$ & $10.0$ \\
         $v$ limits (IRL) & $[2.0, 15.0] m/s$ \\
         $v$ limits (MPC)& $[1.5, 3.5] m/s$ \\
         $\delta$ limits& $[-0.52, 0.52] rad$ \\
         $\omega$ limits& $[-.0.2, 0.2] rad/s$ \\
         $dt$(IRL) & $0.1s$ \\
         $dt$(MPC) & $0.15s$ \\
    \end{tabular}
    \caption{Table of Parameter Values for ATV}
    \label{tab:model_params}
\end{table}

\begin{table}[]
    \centering
    \begin{tabular}{c|c}
        Parameter & Value \\
        \hline
        Iterations / step (IRL) & 10 \\
        Iterations / step (MPC) & 1 \\
        $H$(IRL) & 75 \\
        $H$(MPC) & 60 \\
        $N$(IRL) & 2048 \\
        $N$(MPC) & 512 \\
        $\kappa$(IRL) & 20 \\
        $\kappa$(MPC) & 10 \\
        $\Sigma$ & $diag([1.0, 0.1])$ \\
        $\lambda$ & 20 \\
        $\alpha$ & 0.9 \\
        $K$ & 10.0 \\
        $dK$ & 5.0 \\
    \end{tabular}
    \caption{MPPI parameters}
    \label{tab:mppi_params}
\end{table}

\subsection{Comparison of Dataset Difficulty}

We present a comparison of our dataset to TartanDrive \cite{triest2022tartandrive}. As mentioned in the main paper, the biggest difference in the datasets is the availability of lidar in our IRL dataset. This allows us to generate maps that are $40m$ in the vehicle's forward direction, as opposed to Tartandrive's $10m$. As a result, we are able to collect a dataset that contains much more aggressive driving. We quantify this in Figure \ref{fig:dataset_compare}, where we report the distribution of three quantities associated with driving aggression; speed, yaw rate and integrated change in height. We can observe that on our IRL dataset exceeds TartanDrive on all quantities, especially speed. Furthermore, the distributions in the IRL dataset appear to be more long-tailed.

\begin{figure*}
     \centering
     \begin{subfigure}[t]{\linewidth}
         \centering
         \includegraphics[scale=0.4]{media/IRL media/tartandrive.png}
         \caption{Tartandrive difficulty}
     \end{subfigure}
     \hfill
     \begin{subfigure}[t]{\linewidth}
         \centering
         \includegraphics[scale=0.4]{media/IRL media/irl_dataset.png}
         \caption{IRL dataset difficulty}
     \end{subfigure}
     \hfill
     \caption{Comparison of dataset difficulties between our dataset and TartanDrive. Our IRL dataset has higher mean difficulty and a wider distribution of difficulty, as well.}
     \label{fig:dataset_compare}
\end{figure*}

